\pdfoutput=1
\documentclass[11pt]{article}

\usepackage[]{packages/acl}

\usepackage{times}
\usepackage{latexsym}
\usepackage[T1]{fontenc} 
\usepackage[utf8]{inputenc}
\usepackage{microtype} 

\usepackage[show]{packages/chato-notes} 
\usepackage[pdftex]{graphicx}           
\usepackage{subcaption}                 
\usepackage{booktabs}                   
\usepackage{amsmath}                    
\usepackage{amsfonts}                   
\usepackage{nicefrac}                   
\usepackage{multirow}                   
\usepackage{pifont}                     
\usepackage{enumitem}                   
\usepackage{cleveref}                   
\usepackage{colortbl}                   
\usepackage{orcidlink}                  
\usepackage{listings}                   

\definecolor{shade}{gray}{0.6}

\setlist[itemize,enumerate]{topsep=0pt,itemsep=-1ex,partopsep=1ex,parsep=1ex}
\crefformat{footnote}{#2\footnotemark[#1]#3}
\newcommand{\at}{\texttt{@}}%
\newcommand{\sub}[1]{\small$_{\textsc{#1}}$}

\newcommand{\shade}[1]{\textcolor{shade}{#1}}

\newcommand{\orcid}[1]{\textsuperscript{\normalsize\orcidlink{#1}}}

\lstdefinestyle{CEE}{language=Python, frame=l,  numbers=left,  numbersep=1em,  xleftmargin=2em, basicstyle=\footnotesize}

\title{ColBERT-XM: A Modular Multi-Vector Representation Model for Zero-Shot Multilingual Information Retrieval}

\author{
Antoine Louis\orcid{https://orcid.org/0000-0001-8392-3852}\textnormal{,} 
Vageesh Saxena\orcid{https://orcid.org/0009-0001-3282-7737}\textnormal{,} 
Gijs van Dijck\orcid{https://orcid.org/0000-0003-4102-4415}\textnormal{,} 
Gerasimos Spanakis\orcid{https://orcid.org/0000-0002-0799-0241} \\
Maastricht University, Netherlands \\
{\small \texttt{\{a.louis, v.saxena, gijs.vandijck, jerry.spanakis\}@maastrichtuniversity.nl}} \\
}

\begin{document}
\maketitle

\begin{abstract}
State-of-the-art neural retrievers predominantly focus on high-resource languages like English, which impedes their adoption in retrieval scenarios involving other languages. Current approaches circumvent the lack of high-quality labeled data in non-English languages by leveraging multilingual pretrained language models capable of cross-lingual transfer. However, these models require substantial task-specific fine-tuning across multiple languages, often perform poorly in languages with minimal representation in the pretraining corpus, and struggle to incorporate new languages after the pretraining phase. In this work, we present a novel modular dense retrieval model that learns from the rich data of a single high-resource language and effectively zero-shot transfers to a wide array of languages, thereby eliminating the need for language-specific labeled data. Our model, ColBERT-XM, demonstrates competitive performance against existing state-of-the-art multilingual retrievers trained on more extensive datasets in various languages. Further analysis reveals that our modular approach is highly data-efficient, effectively adapts to out-of-distribution data, and significantly reduces energy consumption and carbon emissions. By demonstrating its proficiency in zero-shot scenarios, ColBERT-XM marks a shift towards more sustainable and inclusive retrieval systems, enabling effective information accessibility in numerous languages. We publicly release our code and models for the community.
\end{abstract}

\begin{figure*}[t]
    \centering
    \includegraphics[width=1.0\linewidth]{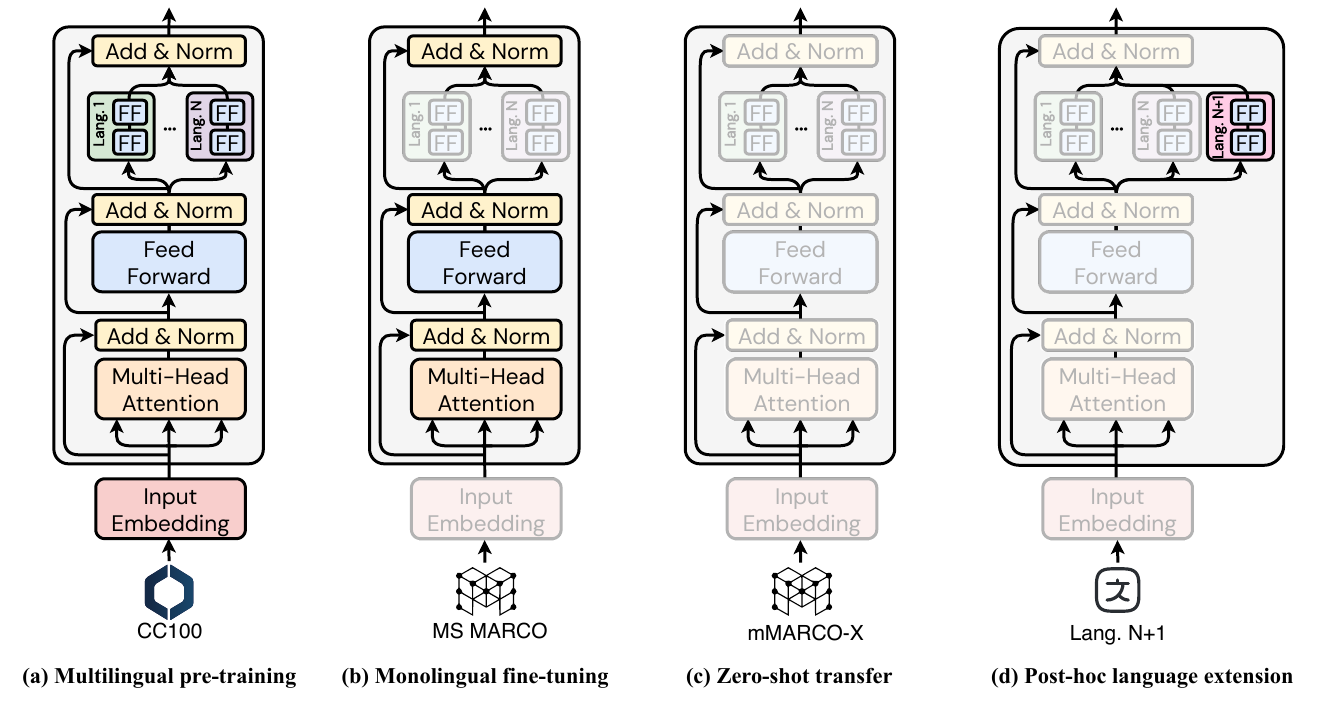}
    \caption{An illustration of ColBERT-XM's modular architecture during its successive learning stages. Components that are blurred indicate they remain frozen throughout the learning phase. \textbf{(a)} First, the model learns language-specific modular adapters at each transformer layer through MLM pretraining on a large multilingual corpus. \textbf{(b)} Next, the model is adapted to the downstream task by fine-tuning its shared weights on the source language while keeping the modular adapters and the embedding layer frozen. \textbf{(c)} The model is then used in a zero-shot fashion by routing the target language's input text through the corresponding modular units. \textbf{(d)} Finally, extra languages can be added post-hoc by learning new modular components only through lightweight MLM training on the new language.}
    \label{fig:learning}
\end{figure*}

\section{Introduction \label{sec:introduction}}
Text retrieval models are integral to various day-to-day applications, including search, recommendation, summarization, and question answering. In recent years, transformer-based models have monopolized textual information retrieval and led to significant progress in the field \citep{lin2021pretrained}. However, the existing literature mostly focuses on improving retrieval effectiveness in a handful of widely spoken languages -- notably English \citep{muennighoff2023mteb} and Chinese \citep{xiao2023cpack} -- whereas other languages receive limited attention.

As a solution, a few studies have suggested fine-tuning multilingual transformer-based encoders, such as mBERT \citep{devlin2019bert}, on aggregated retrieval data across various languages. Nonetheless, this approach faces two major challenges. First, acquiring high-quality relevance labels for various languages proves difficult, particularly for those with fewer resources. Consequently, languages with insufficient representation in training data experience a proficiency gap compared to widely represented ones \citep{macavaney2020teaching}. Second, when these multilingual transformers are pretrained on too many languages, their performance on downstream tasks worsens. This issue, known as the \textsl{curse of multilinguality} \citep{conneau2020unsupervised}, underscores the challenge of developing models that effectively accommodate a broader spectrum of languages.

Our work addresses the challenges above by introducing ColBERT-XM, a novel multilingual dense retrieval model built upon the recent XMOD architecture \citep{pfeiffer2022lifting}, which combines shared and language-specific parameters pretrained from the start to support the following key features:
\begin{enumerate}
    \item \textit{Reduced dependency on multilingual data}: Our XMOD-based retriever is designed to learn through monolingual fine-tuning, capitalizing on the rich data from a high-resource language, like English, thereby reducing the need for extensive multilingual datasets.
    \item \textit{Zero-shot transfer across languages}: Despite being fine-tuned in a single language, our retriever's modular components enable effective knowledge transfer to a variety of underrepresented languages without any further training.
    \item \textit{Post-hoc language addition}: Unlike conventional multilingual models, our modular retriever is easily extendable to languages unseen during pretraining, while mitigating the curse of multilinguality.
\end{enumerate}

Practically, ColBERT-XM learns to effectively predict relevance between queries and passages using only a limited set of English examples, leveraging the late interaction approach introduced in ColBERT \citep{khattab2020colbert}. Our experimental results demonstrate competitive performance across a diverse range of languages against state-of-the-art multilingual models trained on vastly larger datasets and many more languages. Moreover, our analysis shows that ColBERT-XM is highly data efficient, as more training data from the same distribution does not markedly enhance its performance. Even so, further investigations reveal our model's strong ability to generalize to out-of-distribution data, despite its limited training. We also provide evidence that multi-vector representations outperform single-vector approaches within our framework. Finally, we underscore our model's sustainability by examining its environmental impact in comparison to established dense retrievers.

In summary, the contributions of this research are threefold. First, we introduce a novel modular dense retriever that, despite being trained exclusively in one language, demonstrates remarkable adaptability to a broad spectrum of languages in a zero-shot configuration. Second, through comprehensive experiments, we compare the effectiveness of employing multi-vector over single-vector representations, explore the influence of the volume of training examples on the overall model's performance, investigate the model's ability to adapt to out-of-distribution data and languages it has not previously encountered, including low-resource ones, and highlight its sustainable environmental footprint. Finally, we release our source code and model checkpoints at {\small \url{https://github.com/ant-louis/xm-retrievers}}.

\section{Related Work \label{sec:related-work}}
\subsection{Multilingual Information Retrieval}
The term ``multilingual'' typically encompasses a wide range of retrieval tasks using one or more languages \citep{hull1996querying}. In our study, we define it as performing monolingual retrieval across multiple languages. 

Monolingual text retrieval approaches have relied on simple statistical metrics based on term frequency, such as TF-IDF and BM25 \citep{robertson1994okapi}, to represent texts and match documents against a given query. With the advent of transformer-based language models, contextualized representations rapidly got incorporated into retrieval models and gave rise to various neural-based retrieval techniques, including cross-encoder models such as monoBERT \citep{nogueira2019multi} and monoT5 \citep{nogueira2020document}, single-vector bi-encoders like DPR \citep{karpukhin2020dense} and ANCE \citep{xiong2021approximate}, multi-vector bi-encoders like ColBERT \citep{khattab2020colbert} and XTR \citep{lee2023rethinking}, and sparse neural models such as uniCOIL \citep{lin2021few} and SPLADE \citep{formal2021splade}. 

Nevertheless, prior work on neural retrievers has predominantly focused on English due to the abundance of labeled training data. In non-English settings, multilingual pretrained language models such as XLM-R \citep{conneau2020unsupervised} and mBERT \citep{devlin2019bert} emerged as an effective solution, capable of adapting the retrieval task across many languages using a shared model \citep{lawrie2023neural}. However, these models proved to suffer from the curse of multilinguality \citep{chang2023when}, have shown substantially reduced monolingual abilities for low-resource languages with smaller pretraining data \citep{wu2020are}, and do not effectively extend to unseen languages after the pretraining phase \citep{pfeiffer2022lifting}.

\subsection{Modular Transformers}
Traditionally, adapting pretrained transformer-based language models to new data settings involves fully fine-tuning all pretrained weights on relevant data. While effective, this process is computationally expensive. As a parameter-efficient alternative, recent works have proposed inserting lightweight ``expert'' modules after each transformer layer \citep{houlsby2019parameter} to capture specific modeling aspects, such as language-specific \citep{pfeiffer2020madx,ansell2021madg} or task-specific \citep{bapna2019simple,he2021on} knowledge. These modular components, commonly referred to as adapters \citep{rebuffi2017learning}, are selectively fine-tuned for the downstream task, the core transformer parameters remaining frozen. 

Despite their growing use in NLP, adapter-based approaches remain relatively untouched in multilingual information retrieval, with existing IR research primarily concentrating on cross-language retrieval \citep{litschko2022parameter, yang2022parameter}, which aims to return documents in a language different from the query. A key limitation of these works is that the additional capacity introduced by adapters \textsl{after} pretraining is not able to mitigate the curse of multilinguality that has already had a catastrophic impact on the shared transformer weights \citep{pfeiffer2022lifting}. In contrast, our method employs a model inherently designed for modularity that learns language-specific capacity \textsl{during} pretraining, effectively avoiding this limitation.

\section{Method \label{sec:method}}
We present a novel multilingual dense retriever that learns to predict relevance between queries and passages via monolingual fine-tuning, while adapting to various languages in a zero-shot configuration. Our model, ColBERT-XM, adopts a traditional bi-encoder architecture (\S\ref{subsec:biencoder}) based on a modular multilingual text encoder (\S\ref{subsec:xmod}), and employs the MaxSim-based late interaction mechanism (\S\ref{subsec:maxsim}) for relevance assessment. The model is optimized through a contrastive learning strategy (\S\ref{subsec:supervision}), and uses a residual compression approach to significantly reduce the space footprint of indexes utilized for fast vector-similarity search at inference time (\S\ref{subsec:inference}). We describe each part in detail below.

\subsection{Bi-Encoder Architecture\label{subsec:biencoder}}
To predict relevance between query $q$ and passage $p$, ColBERT-XM uses the popular bi-encoder architecture \citep{gillick2018end}, which consists of two learnable text encoding functions $f(\cdot; \boldsymbol{\gamma}_i): \mathcal{W}^{n} \mapsto \mathbb{R}^{n \times d}$, parameterized by $\boldsymbol{\gamma}_i$, that map input text sequences of $n$ terms from vocabulary $\mathcal{W}$ to $d$-dimensional real-valued term vectors, i.e.,
\begin{equation}
\label{eq:1}
\begin{split}
    \mathbf{\hat{H}}_q &= f\!\left([q_1,q_2,\cdots, q_i];\boldsymbol{\gamma}_1 \right), \text{and}\\
    \mathbf{\hat{H}}_p &=f\!\left([p_1,p_2,\cdots, p_j];\boldsymbol{\gamma}_2 \right).
\end{split}
\end{equation}
The main idea behind this architecture is to find values for parameters $\boldsymbol{\gamma}_i$ such that a straightforward similarity function $\mathrm{sim}: \mathbb{R}^{n \times d} \times \mathbb{R}^{m \times d} \mapsto \mathbb{R}_+$ approximates the semantic relevance between $q$ and $p$ by operating on their bags of contextualized term embeddings, i.e.,
\begin{equation}
\operatorname{score}\!\left(q, p\right)=\operatorname{sim}\!\left(\mathbf{\hat{H}}_q, \mathbf{\hat{H}}_p\right).
\end{equation}
This scoring approach, known as \textsl{late interaction} \citep{khattab2020colbert}, as interactions between the query and passage are delayed after their independent encoding computations, stands out for its computational efficiency \citep{reimers2019sentence}. This contrasts with the popular cross-encoder architecture \citep{nogueira2019multi}, which encodes the queries and passages jointly to learn rich interactions directly within the model.

In this work, we use a \textsl{siamese} bi-encoder, where queries and passages are encoded by two identical copies of a shared network (i.e., $\boldsymbol{\gamma}_1 = \boldsymbol{\gamma}_2$).

\begin{figure}[t]
    \centering
    \includegraphics[width=0.85\linewidth]{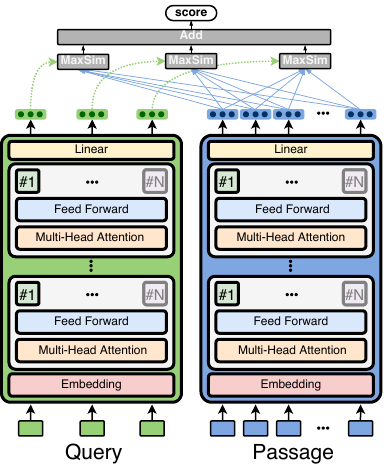}
    \caption{Illustration of the multi-vector late interaction paradigm used in our proposed ColBERT-XM model.}
    \label{fig:late_interaction}
\end{figure}

\subsection{Modular Language Representation\label{subsec:xmod}}
To overcome the limitations posed by multilingual transformer-based encoders outlined in \Cref{sec:introduction}, we use the XMOD model \citep{pfeiffer2022lifting} as our backbone text encoder. As depicted in \Cref{fig:learning}a, XMOD extends the transformer architecture by incorporating language-specific adapters \citep{houlsby2019parameter} at every transformer layer, which are learned from the start \textsl{during} the masked language modeling (MLM) pretraining phase. This method contrasts with conventional adapter-based approaches that typically extend pretrained multilingual models post-pretraining, thereby building upon sub-optimal parameter initialization already affected by the curse of multilinguality.

Formally, our modular language representation model is defined as a learnable encoding function $g(\cdot; \boldsymbol{\theta}, \boldsymbol{\phi}_i): (\mathcal{W}^{k}, \mathcal{L}) \mapsto \mathbb{R}^{k \times d}$, with shared parameters $\boldsymbol{\theta}$ and language-specific parameters $\boldsymbol{\phi}_i$, that maps a text sequence $t$ of $k$ terms from vocabulary $\mathcal{W}$ in language $\mathcal{L}_i$ to $d$-dimensional real-valued representations. Let $\mathbf{W}_{\mathrm{out}} \in \mathbb{R}^{d \times d_{\mathrm{out}}}$ be a linear layer with no activations designed to compress the dimensions of the output representation vectors, \Cref{eq:1} then becomes
\begin{equation}
\begin{split}
    \mathbf{\hat{H}}_t &= g\!\left([t_1,\cdots, t_k];\boldsymbol{\theta},\boldsymbol{\phi}_i\right) \cdot \mathbf{W}_{\mathrm{out}} \\ 
    &= \left[\boldsymbol{\hat{h}}_{1}^{t}, \boldsymbol{\hat{h}}_{2}^{t}, \cdots, \boldsymbol{\hat{h}}_{k}^{t}\right].
\end{split}
\end{equation}
A key benefit of employing XMOD over traditional multilingual transformers is its proven adaptability to accommodate new languages after the initial pretraining phase while maintaining performance across previously included languages, thereby effectively counteracting the curse of multilinguality. Furthermore, \citet{pfeiffer2022lifting} demonstrated that the per-language performance remains consistent whether a language is included during pretraining or added afterward. This suggests that XMOD can potentially encompass numerous languages by pretraining on a subset of languages for which sufficient text data exists, and subsequently adapting to additional, underrepresented languages without deteriorating overall performance. As illustrated in \Cref{fig:learning}d, the post-hoc inclusion of a new language involves learning additional language-specific modular components only through lightweight MLM training on the new language.

\subsection{MaxSim-based Late Interaction\label{subsec:maxsim}}
ColBERT-XM adopts the fine-granular late interaction scoring mechanism of ColBERT, depicted in \Cref{fig:late_interaction}. This mechanism calculates the cosine similarity across all pairs of query and passage embeddings, applies max-pooling across the resulting similarity scores for each query term, and then sum the maximum values across query terms to derive the overall relevance estimate, i.e.,
\begin{equation}
\operatorname{sim}\!\left(\mathbf{\hat{H}}_{\tilde{q}}, \mathbf{\hat{H}}_{\tilde{p}}\right)=\sum_{i=1}^{n} \max_{j=1}^{m}\ \mathrm{cos}\!\left(\boldsymbol{\hat{h}}_{i}^{\tilde{q}}, \boldsymbol{\hat{h}}_{j}^{\tilde{p}} \right),
\end{equation}
where $\tilde{q}$ and $\tilde{p}$ correspond to sequences obtained after incorporating special tokens into $q$ and $p$, respectively, and truncating to preset maximum lengths $n$ and $m$. More specifically, we have
\begin{equation}
\begin{split}
    \tilde{p} &= \left[\textsc{[cls]}, \textsc{[p]}, p_1, \cdots, p_j \right], \text{and}\\
    \tilde{q} &= \left[\textsc{[cls]}, \textsc{[q]}, q_1, \cdots, q_i, \textsc{[m]}, \cdots, \textsc{[m]} \right], 
\end{split}
\end{equation}
where \textsc{[m]} is a mask token appended to queries to reach the predefined length $n$. This padding strategy serves as a query augmentation technique, enhancing the model's ability to interpret short queries through the generation of extra contextualized embeddings at the mask positions. The special tokens \textsc{[p]} and \textsc{[q]} enable the shared XMOD-based encoder to differentiate between passage and query input sequences, respectively.

\subsection{Supervision\label{subsec:supervision}}
\label{sec:supervision}
Let $\mathcal{B}=\{(q_i, p^{+}_i, p_{\textsc{h},i}^{-})\}_{i=1}^{N}$ be a batch of $N$ training instances, each comprising a query $q_i$ associated with a positive passage $p^{+}_i$ and a hard negative passage $p_{\textsc{h},i}^{-}$. By considering the passages paired with all other queries within the same batch, we can enrich each training triple with an additional set of 2($N-$1) \textsl{in-batch} negatives $\mathcal{P}_{\textsc{ib},i}^{-} = \{p^{+}_j,p_{\textsc{h},j}^{-}\}_{j \neq i}^{N}$. Given these augmented training samples, we optimize our model using a contrastive learning strategy that combines two established ranking loss functions, expressed as
\begin{equation}
    \mathcal{L}_{\textsc{total}}\left(q_i, p^{+}_i, p_{\textsc{h},i}^{-}, \mathcal{P}_{\textsc{ib},i}^{-}\right) = \mathcal{L}_{\textsc{pair}}
    + \mathcal{L}_{\textsc{ib}}
\end{equation}
where $\mathcal{L}_{\textsc{pair}}$ is the pairwise softmax cross-entropy loss computed over predicted scores for the positive and hard negative passages, used in ColBERTv1 \citep{khattab2020colbert} and defined as
\begin{equation}
\label{eq:loss_pair}
    \mathcal{L}_{\textsc{pair}} = -\log \frac{e^{\operatorname{score}(q_i,p^{+}_i)}}
{e^{\operatorname{score}(q_i,p^{+}_i)} + e^{\operatorname{score}(q_i,p_{\textsc{h},i}^{-})}},
\end{equation}
while $\mathcal{L}_{\textsc{ib}}$ is the in-batch sampled softmax cross-entropy loss added as an enhancement for optimizing ColBERTv2 \citep{santhanam2022colbertv2}:
\begin{equation}
\label{eq:loss_inbatch}
    \mathcal{L}_{\textsc{ib}} = -\log \frac{e^{\operatorname{score}(q_i,p^{+}_i)}}
{\sum_{p \in \mathcal{P}_{\textsc{ib},i}^{-} \cup \left\{p^{+}_i,p_{\textsc{h},i}^{-}\right\}} e^{\operatorname{score}(q_i, p)}}.
\end{equation}
These contrastive losses aim to learn a high-quality embedding function so that relevant query-passage pairs achieve higher similarity than irrelevant ones.

\begin{table*}[t]
\centering
\resizebox{\textwidth}{!}{%
\begin{tabular}{ll|ccc|cccccccccccccc|c}
\toprule
& \multirow{2}{*}{\textbf{Model}} & \textbf{\# Training} & \textbf{\# Training} & \textbf{\# Active} & \multirow{2}{*}{\textbf{\textit{en}}} & \multirow{2}{*}{\textbf{\textit{es}}} & \multirow{2}{*}{\textbf{\textit{fr}}} & \multirow{2}{*}{\textbf{\textit{it}}} & \multirow{2}{*}{\textbf{\textit{pt}}} & \multirow{2}{*}{\textbf{\textit{id}}} & \multirow{2}{*}{\textbf{\textit{de}}} & \multirow{2}{*}{\textbf{\textit{ru}}} & \multirow{2}{*}{\textbf{\textit{zh}}} & \multirow{2}{*}{\textbf{\textit{ja}}} & \multirow{2}{*}{\textbf{\textit{nl}}} & \multirow{2}{*}{\textbf{\textit{vi}}} & \multirow{2}{*}{\textbf{\textit{hi}}} & \multirow{2}{*}{\textbf{\textit{ar}}} & \multirow{2}{*}{\textbf{Avg}} \\
&&\textbf{Examples}&\textbf{Languages}&\textbf{Params}&&&&&&&&&&&&&&&\\
\midrule
\multicolumn{2}{l}{\textbf{Lexical systems}} & \\
\shade{1} & BM25 (Pyserini) & - & - & - & 18.4 & 15.8 & 15.5 & 15.3 & 15.2 & 14.9 & 13.6 & 12.4 & 11.6 & 14.1 & 14.0 & 13.6 & 13.4 & 11.1 & 14.2 \\
\multicolumn{2}{l}{\textbf{Cross-encoders}} & \\
\shade{2} & mT5{\sub{base}} \citep{bonifacio2021mmarco} & 12.8M & 9 & 390M & \cellcolor{RedOrange!15}36.6 & \cellcolor{RedOrange!15}31.4 & \cellcolor{RedOrange!15}30.2 & \cellcolor{RedOrange!15}30.3 & \cellcolor{RedOrange!15}30.2 & \cellcolor{RedOrange!15}29.8 & \cellcolor{RedOrange!15}28.9 & \cellcolor{RedOrange!15}26.3 & \cellcolor{RedOrange!15}24.9 & \cellcolor{RoyalBlue!15}26.7 & \cellcolor{RoyalBlue!15}29.2 & \cellcolor{RoyalBlue!15}25.6 & \cellcolor{RoyalBlue!15}26.6 & \cellcolor{RoyalBlue!15}23.5 & 28.6\\
\shade{3} & mMiniLM \citep{bonifacio2021mmarco} & 80.0M & 9 & 107M & \cellcolor{RedOrange!15}36.6 & \cellcolor{RedOrange!15}30.9 & \cellcolor{RedOrange!15}29.6 & \cellcolor{RedOrange!15}29.1 & \cellcolor{RedOrange!15}28.9 & \cellcolor{RedOrange!15}29.3 & \cellcolor{RedOrange!15}27.8 & \cellcolor{RedOrange!15}25.1 & \cellcolor{RedOrange!15}24.9 & \cellcolor{RoyalBlue!15}26.3 & \cellcolor{RoyalBlue!15}27.6 & \cellcolor{RoyalBlue!15}24.7 & \cellcolor{RoyalBlue!15}26.2 & \cellcolor{RoyalBlue!15}21.9 & 27.8 \\
\multicolumn{2}{l}{\textbf{Dense single-vector bi-encoders}} & \\
\shade{4} & DPR-X \citep{yang2022c3} & 25.6M & 4 & 550M & \cellcolor{RedOrange!15}24.5 & \cellcolor{RoyalBlue!15}19.6 & \cellcolor{RoyalBlue!15}18.9 & \cellcolor{RoyalBlue!15}18.3 & \cellcolor{RoyalBlue!15}19.0 & \cellcolor{RoyalBlue!15}16.9 & \cellcolor{RoyalBlue!15}18.2 & \cellcolor{RedOrange!15}17.7 & \cellcolor{RedOrange!15}14.8 & \cellcolor{RoyalBlue!15}15.4 & \cellcolor{RoyalBlue!15}18.5 & \cellcolor{RoyalBlue!15}15.1 & \cellcolor{RoyalBlue!15}15.4 & \cellcolor{RoyalBlue!15}12.9 & 17.5\\
\shade{5} & mE5{\sub{base}} \citep{wang2022text} & 5.1B & 16 & 278M & \cellcolor{RedOrange!15}35.0 & \cellcolor{RedOrange!15}28.9 & \cellcolor{RedOrange!15}30.3 & \cellcolor{RedOrange!15}28.0 & \cellcolor{RedOrange!15}27.5 & \cellcolor{RedOrange!15}26.1 & \cellcolor{RedOrange!15}27.1 & \cellcolor{RedOrange!15}24.5 & \cellcolor{RedOrange!15}22.9 & \cellcolor{RedOrange!15}25.0 & \cellcolor{RedOrange!15}27.3 & \cellcolor{RedOrange!15}23.9 & \cellcolor{RedOrange!15}24.2 & \cellcolor{RedOrange!15}20.5 & 26.5 \\
\multicolumn{2}{l}{\textbf{Dense multi-vector bi-encoders}} & \\
\shade{6} & mColBERT \citep{bonifacio2021mmarco} & 25.6M & 9 & 180M & \cellcolor{RedOrange!15}35.2 & \cellcolor{RedOrange!15}30.1 & \cellcolor{RedOrange!15}28.9 & \cellcolor{RedOrange!15}29.2 & \cellcolor{RedOrange!15}29.2 & \cellcolor{RedOrange!15}27.5 & \cellcolor{RedOrange!15}28.1 & \cellcolor{RedOrange!15}25.0 & \cellcolor{RedOrange!15}24.6 & \cellcolor{RoyalBlue!15}23.6 & \cellcolor{RoyalBlue!15}27.3 & \cellcolor{RoyalBlue!15}18.0 & \cellcolor{RoyalBlue!15}23.2 & \cellcolor{RoyalBlue!15}20.9 & 26.5\\
\midrule
\multicolumn{2}{l}{\textbf{Ours}} &\\
\shade{7} & ColBERT-XM & 6.4M & 1 & 277M & \cellcolor{RedOrange!15}37.2 & \cellcolor{RoyalBlue!15}28.5 & \cellcolor{RoyalBlue!15}26.9 & \cellcolor{RoyalBlue!15}26.5 & \cellcolor{RoyalBlue!15}27.6 & \cellcolor{RoyalBlue!15}26.3 & \cellcolor{RoyalBlue!15}27.0 & \cellcolor{RoyalBlue!15}25.1 & \cellcolor{RoyalBlue!15}24.6 & \cellcolor{RoyalBlue!15}24.1 & \cellcolor{RoyalBlue!15}27.5 & \cellcolor{RoyalBlue!15}22.6 & \cellcolor{RoyalBlue!15}23.8 & \cellcolor{RoyalBlue!15}19.5 & 26.2 \\
\bottomrule
\end{tabular}
}
\caption{MRR\at10 results on mMARCO small dev set. Performance on languages encountered during fine-tuning is highlighted in \colorbox{RedOrange!15}{orange}, whereas zero-shot performance is highlighted in \colorbox{RoyalBlue!15}{blue}. ColBERT-XM reaches near state-of-the-art results while trained on one language only with much fewer examples than competitive models. }
\label{tab:results}
\end{table*}

\subsection{Inference\label{subsec:inference}}
Since passages and queries are encoded independently, passage embeddings can be precomputed and indexed offline through efficient vector-similarity search data structures, using the \href{https://github.com/facebookresearch/faiss}{\texttt{faiss}} library \citep{jonhson2021billion}. Instead of directly indexing the passage representations as in ColBERTv1, which requires substantial storage even when compressed to 32 or 16 bits, we adopt the centroid-based indexing approach introduced in ColBERTv2, 
as detailed in \Cref{app:indexing}.

\section{Experiments \label{sec:experiments}}

\subsection{Experimental Setup \label{subsec:setup}}
\paragraph{Data.}
For training, we follow ColBERTv1 and use triples from the MS MARCO passage ranking dataset \citep{nguyen2018msmarco}, which contains 8.8M passages and 539K training queries. However, unlike the original work that uses the BM25 negatives provided by the official dataset, we sample harder negatives mined from 12 distinct dense retrievers.\footnote{\scriptsize\url{https://huggingface.co/datasets/sentence-transformers/msmarco-hard-negatives}} For a comprehensive evaluation across various languages, we consider the small development sets from mMARCO \citep{bonifacio2021mmarco}, a machine-translated variant of MS MARCO in 13 languages, each comprising 6980 queries. To assess out-of-distribution performance, we use the test sets from Mr. \textsc{TyDi} \citep{zhangtydi2021}, another multilingual open retrieval dataset including low-resource languages not present in mMARCO.

\paragraph{Implementation.} 
We train our model for 50k steps using the AdamW optimizer \citep{loshchilov2017decoupled} with a batch size of 128, a peak learning rate of 3e-6 with warm up along the first 10\% of training steps and linear scheduling. We set the embedding dimension to $d_{\mathrm{out}}$$=$$128$, and fix the maximum sequence lengths for questions and passages at $n$$=$$32$ and $m$$=$$256$, respectively. Training is performed on one 80GB NVIDIA H100 GPU hosted on a server with a dual 20-core Intel Xeon E5-2698 v4 CPU \at 2.20GHz and 512GB of RAM. We use the following Python libraries: \href{https://github.com/huggingface/transformers}{\texttt{transformers}} \citep{wolf2020transformers}, \href{https://github.com/UKPLab/sentence-transformers}{\texttt{sentence-transformers}} \citep{reimers2019sentence}, \href{https://github.com/stanford-futuredata/ColBERT}{\texttt{colbert-ir}} \citep{khattab2020colbert}, and \href{https://github.com/wandb/wandb}{\texttt{wandb}} \citep{biewald2020wandb}.

\paragraph{Metrics \& evaluation.}
To measure effectiveness, we use the official metrics for each query set, i.e., mean reciprocal rank at cut-off 10 (MRR\at10) for MS MARCO, and recall at cut-off 100 (R\at100) along MRR\at100 for Mr. \textsc{TyDi}. We compare our model against established multilingual baselines spanning four retrieval methodologies. For lexical matching, we report the widely adopted bag-of-words BM25 function \citep{robertson1994okapi}. For the cross-encoders, we include two classification models based on mMiniLM{\sub{L6}} \citep{wang2021minilm} and mT5{\sub{base}} \citep{xue2021mt5}, each fine-tuned on mMARCO pairs across 9 languages \citep{bonifacio2021mmarco}. The dense single-vector bi-encoders are derived from XLM-R \citep{conneau2020unsupervised} and have been fine-tuned on samples in 4 \citep{yang2022c3} and 16 languages \citep{wang2022text}, respectively. Lastly, we report the performance of a dense multi-vector bi-encoder built on mBERT{\sub{base}} \citep{devlin2019bert} and fine-tuned on mMARCO samples across 9 languages \citep{bonifacio2021mmarco}.

\begin{figure*}[t]
\centering
\begin{subfigure}[t]{.24\textwidth}
  \centering
  \includegraphics[width=\linewidth]{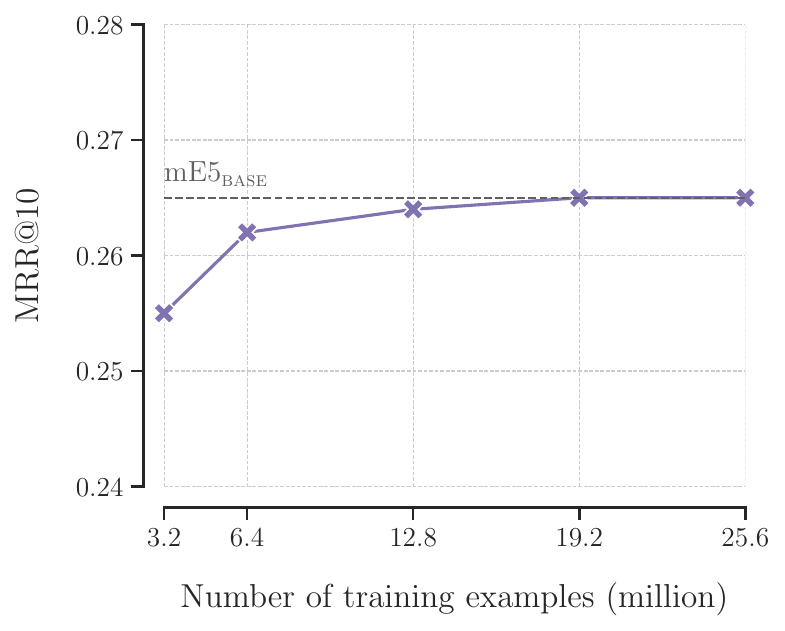}
\end{subfigure}
\begin{subfigure}[t]{.24\textwidth}
  \centering
  \includegraphics[width=\linewidth]{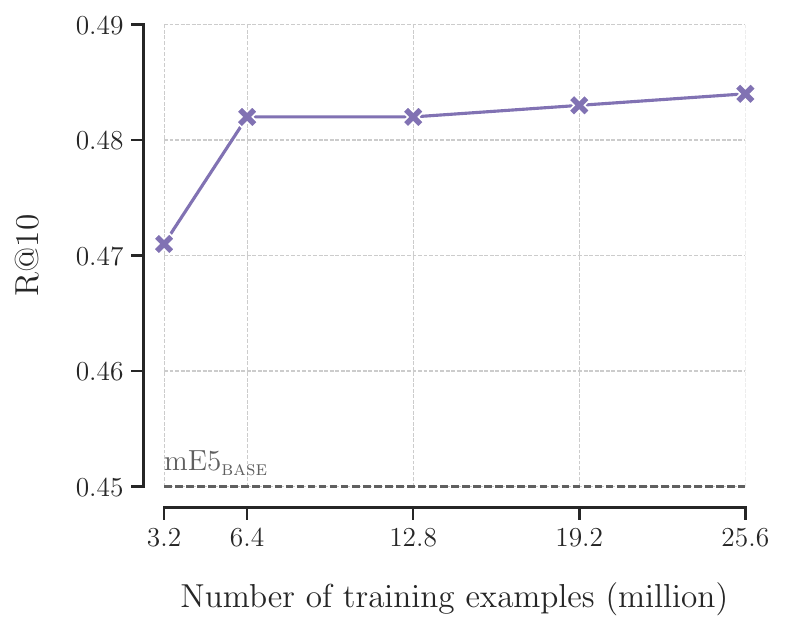}
\end{subfigure}
\begin{subfigure}[t]{.24\textwidth}
  \centering
  \includegraphics[width=\linewidth]{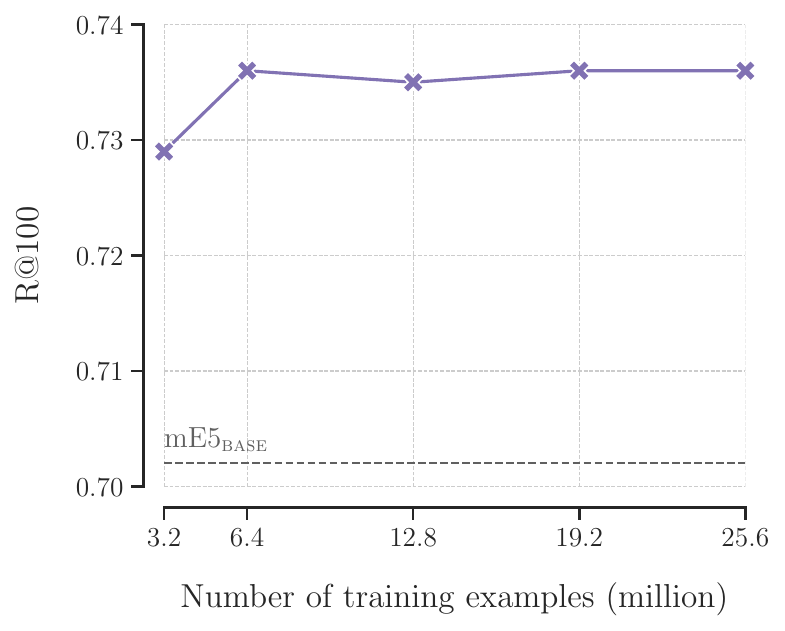}
\end{subfigure}
\begin{subfigure}[t]{.24\textwidth}
  \centering
  \includegraphics[width=\linewidth]{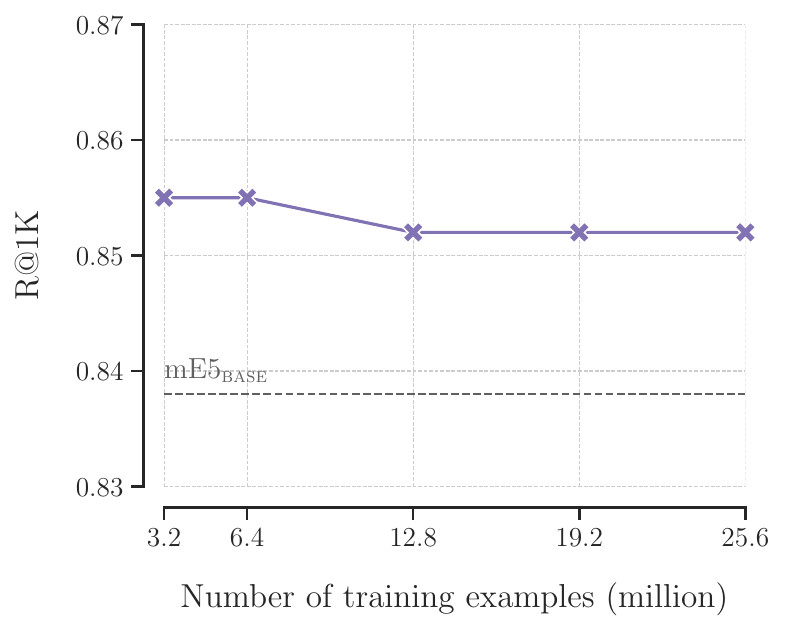}
\end{subfigure}
\caption{Performance of ColBERT-XM on mMARCO small dev set, based on the volume of training examples.}
\label{fig:training_examples}
\end{figure*}

\subsection{Main Results \label{subsec:results}}
\Cref{tab:results} reports results using the official MRR\at10 metric for the 14 languages included in mMARCO. In its training language (i.e. English), ColBERT-XM outperforms all multilingual baselines. The underperformance of certain models, like mT5{\sub{base}} and mColBERT, can partly be attributed to their exposure to fewer English examples given their training across 9 languages with 12.8M and 25.6M samples distributed evenly -- resulting in only 1.4M and 2.8M English examples, respectively, compared to ColBERT-XM's 6.4M training set. Conversely, models such as mMiniLM{\sub{l6}} and mE5{\sub{base}}, despite being exposed to a larger number of English examples, still underperform, suggesting that the modular architecture of ColBERT-XM may offer intrinsic benefits over conventional multilingual models.

In languages on which ColBERT-XM was not trained but the baselines were, we observe comparable performance. For instance, when excluding English, the difference in average performance between our model and mE5{\sub{base}} is merely 0.5\%, even though mE5{\sub{base}} was trained in 15 additional languages and 800,000 times more data samples. In languages on which neither ColBERT-XM nor the baselines were trained, we note a slight enhancement in performance among the computationally expensive cross-encoder models, while both the non-modular single-vector and multi-vector bi-encoders lag behind our model in performance.

Overall, ColBERT-XM demonstrates strong knowledge transfer and generalization capabilities across languages while trained on a significantly smaller monolingual set.

\subsection{Further Analysis \label{subsec:analysis}}
In this section, we conduct a thorough analysis of several key aspects of our proposed methodology, including the influence of greater volumes of training data on ColBERT-XM's performance (\S\ref{subsubsec:more_training_data}), a performance comparison with a modular single-vector representation variant (\S\ref{subsubsec:single_vector_comparison}), the model's ability to generalize to out-of-distribution data (\S\ref{subsubsec:out_of_domain_generalization}), and its environmental footprint compared to existing multilingual retrievers (\S\ref{subsubsec:environment_footprint}).

\subsubsection{How does training on more examples affect ColBERT-XM performance?}
\label{subsubsec:more_training_data}
Despite being trained on substantially fewer examples, ColBERT-XM demonstrates competitive results compared to existing multilingual models, raising the question of whether an increased volume of training data would further enhance its performance. To investigate, we train five instances of our modular retriever on a varying number of MS MARCO training triples, namely 3.2M, 6.4M, 12.8M, 19.2M, and 25.6M examples. \Cref{fig:training_examples} shows the resulting models' performance on the mMARCO small dev set across MRR\at10 and recall at various cut-offs, alongside the fixed performance of mE5{\sub{base}} for comparison. The results reveal an initial performance boost with an increase in training data, which plateaus quickly after 6.4M examples, suggesting diminishing returns from additional data of the same distribution. This contrasts with existing baselines that were trained on comparatively more samples from diverse languages to reach their peak performance, thereby underscoring ColBERT-XM's efficiency in low-resource scenarios. For a comprehensive breakdown of performance across individual languages, we refer to \Cref{tab:training_samples} in \Cref{app:experiment_details}.

\begin{table*}[t]
\centering
\resizebox{\textwidth}{!}{%
\begin{tabular}{llc|llllllllll|c}
\toprule
&\textbf{Model} & \textbf{Type} & \multicolumn{1}{c}{\textbf{\textit{ar}}} & \multicolumn{1}{c}{\textbf{\textit{bn}}} & \multicolumn{1}{c}{\textbf{\textit{en}}} & \multicolumn{1}{c}{\textbf{\textit{fi}}} & \multicolumn{1}{c}{\textbf{\textit{id}}} & \multicolumn{1}{c}{\textbf{\textit{ja}}} & \multicolumn{1}{c}{\textbf{\textit{ko}}} & \multicolumn{1}{c}{\textbf{\textit{ru}}} & \multicolumn{1}{c}{\textbf{\textit{sw}}} & \multicolumn{1}{c|}{\textbf{\textit{te}}} & \textbf{Avg} \\ 
\midrule\midrule
\multicolumn{3}{c}{}&\multicolumn{10}{c}{\textbf{MRR\at100}} & \\
\shade{1} & BM25 (Pyserini) & \textsc{lexical} & 36.8 & 41.8 & 14.0 & 28.4 & 37.6 & 21.1 & 28.5 & 31.3 & 38.9 & 34.3 & 31.3 \\
\shade{2} & mT5{\sub{base}} \citep{bonifacio2021mmarco} & \textsc{cross} & \textbf{62.2} & \textbf{65.1} & 35.7$^{\dagger}$ & \textbf{49.5} & 61.1$^{\dagger}$ & \textbf{48.1} & \textbf{47.4} & \textbf{52.6}$^{\dagger}$ & \textbf{62.9} & \textbf{66.6} & \textbf{55.1} \\
\shade{3} & mColBERT \citep{bonifacio2021mmarco} & \textsc{multi} & \underline{55.3} & 48.8 & 32.9$^{\dagger}$ & 41.3 & 55.5$^{\dagger}$ & 36.6 & 36.7 & 48.2$^{\dagger}$ & 44.8 & \underline{61.6} & 46.1 \\
\shade{4} & ColBERT-XM (ours) & \textsc{multi} & 55.2 & \underline{56.6} & \textbf{36.0}$^{\dagger}$ & \underline{41.8} & \underline{57.1} & \underline{42.1} & \underline{41.3} & \underline{52.2} & \underline{56.8} & 50.6 & \underline{49.0} \\
\midrule
\multicolumn{3}{c}{}&\multicolumn{10}{c}{\textbf{R\at100}} & \\
\shade{5} & BM25 (Pyserini) & \textsc{lexical} & 79.3 & 86.9 & 53.7 & 71.9 & 84.3 & 64.5 & 61.9 & 64.8 & 76.4 & 75.8 & 72.0 \\
\shade{6} & mT5{\sub{base}} \citep{bonifacio2021mmarco} & \textsc{cross} & \underline{88.4} & \textbf{92.3} & 72.4$^{\dagger}$ & \textbf{85.1} & 92.8$^{\dagger}$ & \underline{83.2} & \underline{76.5} & 76.3$^{\dagger}$ & \underline{83.8} & 85.0 & 83.5 \\
\shade{7} & mColBERT \citep{bonifacio2021mmarco} & \textsc{multi} & 85.9 & \underline{91.8} & 78.6$^{\dagger}$ & 82.6 & 91.1$^{\dagger}$ & 70.9 & 72.9 & \underline{86.1}$^{\dagger}$ & 80.8 & \textbf{96.9} & \underline{83.7} \\
\shade{8} & ColBERT-XM (ours) & \textsc{multi} & \textbf{89.6} & 91.4 & \textbf{83.7}$^{\dagger}$ & \underline{84.4} & \textbf{93.8} & \textbf{84.9} & \textbf{77.6} & \textbf{89.1} & \textbf{87.1} & \underline{93.3} & \textbf{87.5} \\
\bottomrule
\end{tabular}
}
\caption{Out-of-domain retrieval performance on Mr. \textsc{TyDi} test set. All supervised models were fine-tuned on one or more languages from mMARCO. $\dagger$ indicates performance on languages encountered during fine-tuning. The best results are marked in \textbf{bold}, and the second best are \underline{underlined}.}
\label{tab:out-of-domain}
\end{table*}

\subsubsection{How does a single-vector representation variant compare to ColBERT-XM?}
\label{subsubsec:single_vector_comparison}
To analyze the effects of single-vector vs. multi-vector representations on our model's performance, we implement a variant of our modular dense retriever that maintains the bi-encoder architecture and modular encoder outlined in Sections \ref{subsec:biencoder} and \ref{subsec:xmod}, respectively, yet adopts a different late interaction scoring mechanism that operates on single-vector representations of the input sequences, i.e.,
\begin{equation}
\operatorname{sim}\!\left(\mathbf{\hat{H}}_{q}, \mathbf{\hat{H}}_{p}\right)= \operatorname{cos}\!\left(\operatorname{pool}\!\left(\mathbf{\hat{H}}_{q}\right), \operatorname{pool}\!\left(\mathbf{\hat{H}}_{p}\right) \right),
\end{equation}
where $\operatorname{pool}: \mathbb{R}^{k \times d} \rightarrow \mathbb{R}^{d}$ distills a global representation for the whole text sequence using mean, max, or \textsc{[cls]} pooling on the corresponding bags of contextualized term embeddings. We train this model, dubbed DPR-XM, on 25.6M MS MARCO triples with a batch size of 128 and learning rate warm up along the first 10\% of steps to a maximum value of 2e-5, after which linear decay is applied.

\begin{figure}[t]
    \centering
    \includegraphics[width=0.5\linewidth]{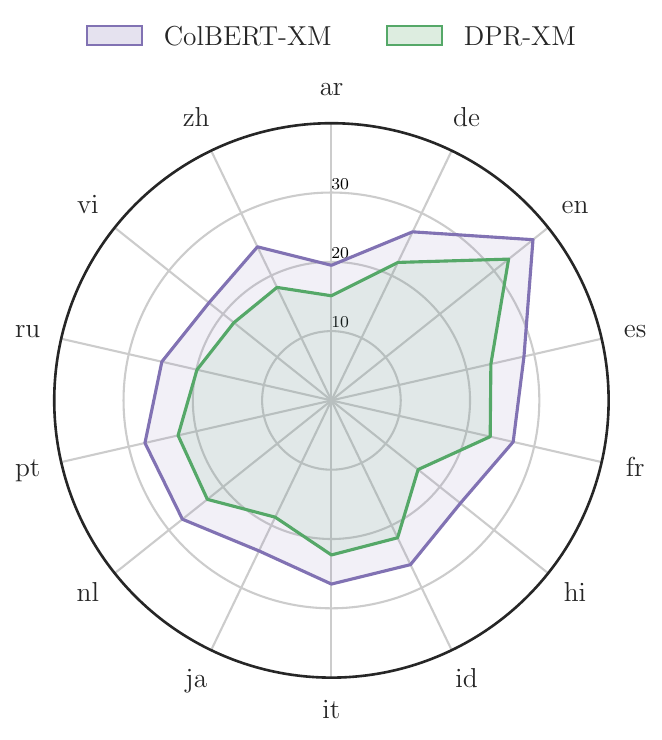}
    \caption{MRR\at10 results of our multi-vector representation retriever (ColBERT-XM) compared to its single-vector counterpart (DPR-XM) on mMARCO dev set.}
    \label{fig:multi_vs_single}
\end{figure}

\Cref{fig:multi_vs_single} illustrates the comparative performance of our XMOD-based dense retrievers. We observe that ColBERT-XM surpasses DPR-XM in the training language (i.e., English) by 4.5\% on MRR\at10. Furthermore, it consistently outperforms DPR-XM across the other 13 languages not encountered during training by an average of 4.9\%. Supported by findings from \citet{santhanam2022colbertv2}, our results confirm that multi-vector models bypass the restrictive information bottleneck inherent in single-vector models, enabling a richer and more nuanced representation of queries and passages, thereby yielding higher retrieval performance. 


\subsubsection{How does ColBERT-XM generalize to out-of-distribution data?}
\label{subsubsec:out_of_domain_generalization}
To assess ColBERT-XM's capabilities for out-of-distribution generalization, we conduct a zero-shot evaluation on Mr. \textsc{TyDi}, encompassing five languages not covered in mMARCO -- notably Swahili, Bengali, and Telugu, which are commonly identified as low-resource. \Cref{tab:out-of-domain} reports the zero-shot performance of ColBERT-XM alongside the BM25, mT5-based cross-encoder, and mColBERT baselines. We find that ColBERT-XM shows substantial generalization across the out-of-distribution data. While not as effective as the computationally expensive cross-attentional mT5{\sub{base}} re-ranking model on the rank-aware MRR\at100 metrics, ColBERT-XM outperforms its non-modular mColBERT counterpart. Notably, on the rank-unaware R\at100 metrics, ColBERT-XM matches closely and even surpasses the more resource-intensive mColBERT and mT5 retrieval models, which have been trained on many more samples and languages. These findings highlight our model's ability to efficiently adapt to domains and languages beyond its original training scope.

\subsubsection{What is the environmental footprint of ColBERT-XM?}
\label{subsubsec:environment_footprint}
Given the growing concerns over carbon emissions and climate change, the environmental impact of AI models has become a crucial issue. In a quest for achieving ever-increasing performance, many works prioritize effectiveness over efficiency, leading to models whose training requires significant energy consumption often derived from non-renewable resources, thereby exacerbating the global carbon footprint. Our comparative analysis demonstrates that ColBERT-XM exhibits reduced energy consumption and carbon emissions while performing comparably to leading retrieval models, underscoring its economic and environmental advantages.\footnote{Estimations were conducted using the \href{https://mlco2.github.io/impact}{MachineLearning Impact calculator} \citep{lacoste2019quantifying}.} \Cref{tab:carbon_footprint} reveals that ColBERT-XM, trained for 7.5 hours only on private infrastructure with a carbon efficiency of 0.432 kgCO$_2$eq/kWh, utilized only 2.3 kWh of power for a carbon footprint of about 1.01 kgCO2eq, which is approximately the amount of emissions produced by burning 0.5 kg of coal. This contrasts significantly with competing models like mE5, which, despite its high performance, consumed about 100$\times$ more power during training (i.e., 230.4 kWh), emitting carbon emissions equivalent to burning 49.6 kg of coal. For reference, we compute the estimated carbon emissions $E_{\mathrm{c}}$ as
\begin{equation}
    E_{\mathrm{c}} = \overbrace{
    \underbrace{P_{\textsc{tdp}}}_{\substack{\text{Thermal} \\ \text{Design Power}}} \times
    \underbrace{T_{\mathrm{train}}}_{\substack{\text{Training} \\ \text{time}}}
   }^{\text{Power consumption}}
   + \underbrace{C_{\mathrm{eff}}}_{\substack{\text{Carbon} \\ \text{efficiency}}}.
\end{equation}

Our analysis not only highlights the potential for reduced carbon emissions associated with multilingual dense retrievers, but also reflects a deliberate stride toward aligning AI models with the pressing need for environmental sustainability. By demonstrating a comparable performance with a fraction of the energy and carbon output, we hope to set a precedent for future research and development in the field, emphasizing the importance of eco-friendly retrieval systems.


\section{Conclusion \label{sec:conclusion}}
This research presents ColBERT-XM, a multilingual dense retrieval model built upon the XMOD architecture, which effectively learns from monolingual fine-tuning in a high-resource language and performs zero-shot retrieval across multiple languages. Despite being trained solely in English, ColBERT-XM demonstrates competitive performance compared to existing state-of-the-art neural retrievers trained on more extensive datasets in various languages. An in-depth analysis reveals that our modular model learns faster, consumes a fraction of energy, and has a lower carbon footprint than existing multilingual models, thereby balancing its efficacy with environmental sustainability goals. Additionally, ColBERT-XM generalizes on out-of-distribution data and low-resource languages without further training, performing closely or surpassing strong retrievers. We believe that our research can help build effective retrieval systems for many languages while eliminating the need for language-specific labeled data, thus fostering inclusivity and linguistic diversity by helping individuals access information in their native languages.

\definecolor{red1}{HTML}{FF9999}
\definecolor{orange1}{HTML}{FFB266}
\definecolor{orange2}{HTML}{FFCC99}
\definecolor{vert1}{HTML}{BFDBB2}
\begin{table}[t]
\centering
\resizebox{\linewidth}{!}{%
\begin{tabular}{ll|lcc|rr}
\toprule
& \multirow{2}{*}{\textbf{Model}} & \multirow{2}{*}{\textbf{Hardware}} & \textbf{TDP} & \textbf{Training} & \textbf{Power} & \textbf{Emission} \\ 
& & & \textbf{(W)} & \textbf{time (h)} & \textbf{(kWh)} & \textbf{(kgCO$_2$eq)} \\\midrule
\shade{1} & mE5{\sub{base}}   & 32$\times$ V100 & 300 & 24 & \cellcolor{red1!80}230.4 & \cellcolor{red1!80}99.52 \\
\shade{2} & mMiniLM{\sub{l6}} & 1$\times$ A100  & 400 & 50 & \cellcolor{orange1!90}20.0 &  \cellcolor{orange1!90}8.64 \\
\shade{3} & mColBERT         & 1$\times$ V100  & 300 & 36 & \cellcolor{orange2!75}10.8 &  \cellcolor{orange2!75}4.67 \\
\shade{4} & mT5{\sub{base}}   & 1$\times$ TPUv3 & 283 & 27 & \cellcolor{orange2!45}7.6 & \cellcolor{orange2!45}3.30 \\
\midrule
\shade{5} & ColBERT-XM     & 1$\times$ H100  & 310 & 7.5 & \cellcolor{vert1!55}2.3 & \cellcolor{vert1!55}1.01 \\
\bottomrule
\end{tabular}
}
\caption{Power efficiency and carbon footprint of existing multilingual retrieval models.}
\label{tab:carbon_footprint}
\end{table}

\section*{Limitations \label{sec:limitations}}
This section enumerates our work's limitations.

\paragraph{Broader evaluation across diverse datasets.}
While our model's evaluation predominantly relies on the mMARCO dataset \citep{bonifacio2021mmarco}, future investigations could benefit from exploring a broader spectrum of multilingual retrieval datasets, such as MIRACL \citep{zhang2022making}, SWIM-IR \citep{thakur2023leveraging}, and MLDR \citep{chen2024bge}. Additionally, examining the model's proficiency in domain-specific retrieval could offer valuable insights into its adaptability to specialized knowledge areas. Unfortunately, such benchmarks are scarce in multilingual contexts.

\paragraph{Distillation of expressive retrieval models.}
Instead of the standard pairwise cross-entropy loss employed in ColBERTv1, a KL-divergence loss aimed at distilling the scores from a more sophisticated cross-encoder model, as introduced in ColBERTv2, could yield notable performance improvement \citep{santhanam2022colbertv2}. Nevertheless, our estimates suggest this supervision scheme would require approximately 9.3 times more computational time for training on our system, surpassing our current resource allocation. As such, we let this exploration for future work.

\paragraph{Adaptability to cross-lingual retrieval.}
While this study presents a multilingual model designed for information retrieval within the same language, investigating its cross-lingual retrieval capabilities -- i.e., identifying relevant passages in a target language based on queries in a different source language -- represents a compelling direction for future research, especially in light of increasing needs for systems that can transcend language barriers.

\paragraph{Model interpretability.}
Enhancing the interpretability of dense retrieval model predictions is essential for boosting user confidence and ensuring system transparency, particularly given the complex linguistic and cultural nuances present in multilingual contexts. Building on seminal works in the area \citep{sudhi2022bibte,anand2023explainable}, our future efforts will focus on deepening our understanding of ColBERT-XM's decision-making mechanisms through detailed analysis of the model’s interpretability features.

\section*{Broader Impacts \label{sec:impacts}}
In this section, we delve into the ethical considerations, societal implications, and potential risks of our proposed methodology.

\paragraph{Ethical considerations.} Our work mostly leverages the widely recognized MS MARCO dataset, which contains over half a million anonymized queries collected from Bing's search logs, ensuring that our data sourcing practices are ethical and protect individual privacy. By leveraging mMARCO's direct translations, we ensure a fair and unbiased distribution of samples across languages, thereby avoiding the reinforcement of stereotypes. Furthermore, the combination of automated translation and manual labeling of the dataset ensures the reliability and precision of the ground truth data. This approach is essential for reducing label bias, which can arise from human annotators' varying proficiency levels and backgrounds.

\paragraph{Societal implications.} Multilingual retrieval models significantly impact society by reducing language barriers and improving information accessibility for all. Our research aims to foster inclusivity and linguistic diversity, helping non-English speakers and those desiring information in their native languages. By developing models capable of effectively retrieving information in lesser-used languages, we contribute to equitable learning opportunities worldwide, enable businesses to serve a diverse international clientele, and prevent the digital marginalization of linguistic minorities.


\paragraph{Potential misuse.} 
The premature deployment of a modular retrieval system presents a few risks. Notably, flaws or biases acquired during the monolingual fine-tuning phase could be inadvertently propagated to other languages when performing zero-shot transfer, thus perpetuating these malfunctions. More generally, the integrity of the underlying knowledge corpus is crucial, as even an effective system may retrieve relevant yet factually inaccurate content, thus unwittingly spreading misinformation. These concerns underscore the need for interpretability of retrieval model predictions to bolster user trust in such systems, which ColBERT-XM addresses with its interpretable MaxSim-based scoring mechanism.

\section*{Acknowledgments}
This research is partially supported by the Sector Plan Digital Legal Studies of the Dutch Ministry of Education, Culture, and Science. In addition, this research was made possible, in part, using the Data Science Research Infrastructure (DSRI) hosted at Maastricht University.

\bibliography{refs.bib}
\bibliographystyle{packages/acl_natbib}

\appendix

\begin{table*}[th!]
\centering
\resizebox{\textwidth}{!}{%
\begin{tabular}{c|cccccccccccccc|c}
\toprule
\textbf{\# Training}   & \multirow{2}{*}{\textbf{\textit{en}}} & \multirow{2}{*}{\textbf{\textit{es}}} & \multirow{2}{*}{\textbf{\textit{fr}}} & \multirow{2}{*}{\textbf{\textit{it}}} & \multirow{2}{*}{\textbf{\textit{pt}}} & \multirow{2}{*}{\textbf{\textit{id}}} & \multirow{2}{*}{\textbf{\textit{de}}} & \multirow{2}{*}{\textbf{\textit{ru}}} & \multirow{2}{*}{\textbf{\textit{zh}}} & \multirow{2}{*}{\textbf{\textit{ja}}} & \multirow{2}{*}{\textbf{\textit{nl}}} & \multirow{2}{*}{\textbf{\textit{vi}}} & \multirow{2}{*}{\textbf{\textit{hi}}} & \multirow{2}{*}{\textbf{\textit{ar}}} & \multirow{2}{*}{\textbf{Avg}} \\
\textbf{Examples}&&&&&&&&&&&&&&&\\ \midrule
\multicolumn{1}{c}{}&\multicolumn{14}{c}{\textbf{MRR\at10}}& \\
3.2M  & 35.7 & 27.7 & 25.9 & 26.2 & 26.9 & 25.3 & 26.2 & 24.4 & 24.0 & 23.9 & 26.5 & 21.8 & 23.2 & 19.2 & 25.5 \\
6.4M  & 37.2 & 28.5 & 26.9 & 26.5 & 27.6 & 26.3 & 27.0 & 25.1 & 24.6 & 24.1 & 27.5 & 22.6 & 23.8 & 19.5 & 26.2 \\
12.8M & 38.1 & 28.6 & 26.8 & 26.9 & 27.5 & 26.6 & 27.1 & 25.4 & 24.9 & 24.2 & 27.3 & 22.9 & 23.8 & 19.5 & 26.4 \\
19.2M & 38.2 & 28.7 & 26.8 & 26.7 & 27.9 & 26.7 & 27.1 & 25.7 & 25.0 & 24.1 & 27.5 & 23.2 & 23.7 & 19.3 & 26.5 \\
25.6M & 38.0 & 28.4 & 26.7 & 26.8 & 27.8 & 26.6 & 27.1 & 26.0 & 25.2 & 24.2 & 27.5 & 23.2 & 23.8 & 19.6 & 26.5 \\ \midrule
\multicolumn{1}{c}{}&\multicolumn{14}{c}{\textbf{R\at10}}& \\
 3.2M & 63.8 & 50.4 & 48.2 & 47.8 & 49.6 & 46.8 & 48.3 & 46.1 & 44.9 & 44.3 & 49.2 & 41.2 & 43.4 & 35.6 & 47.1 \\
 6.4M & 65.7 & 52.0 & 49.2 & 48.2 & 50.5 & 48.3 & 49.5 & 47.3 & 46.0 & 44.6 & 49.8 & 42.4 & 44.2 & 36.5 & 48.2 \\
12.8M & 66.4 & 51.8 & 48.7 & 48.6 & 50.5 & 48.3 & 49.6 & 47.1 & 45.9 & 45.0 & 50.0 & 42.3 & 43.8 & 36.4 & 48.2 \\
19.2M & 67.0 & 52.0 & 49.1 & 48.2 & 50.4 & 48.9 & 49.6 & 47.8 & 46.0 & 44.8 & 50.0 & 42.8 & 43.6 & 35.7 & 48.3 \\
25.6M & 67.0 & 51.9 & 48.7 & 48.8 & 50.5 & 48.6 & 49.7 & 47.9 & 46.4 & 45.0 & 50.0 & 42.7 & 43.8 & 36.2 & 48.4 \\ \midrule
\multicolumn{1}{c}{}&\multicolumn{14}{c}{\textbf{R\at100}}& \\
 3.2M & 88.5 & 77.2 & 75.1 & 73.6 & 75.3 & 73.3 & 73.1 & 73.0 & 71.6 & 71.2 & 74.4 & 66.8 & 68.6 & 59.3 & 72.9 \\
 6.4M & 89.3 & 77.5 & 75.2 & 74.1 & 75.8 & 74.5 & 73.9 & 73.6 & 72.2 & 71.4 & 75.2 & 67.5 & 69.8 & 60.4 & 73.6 \\
12.8M & 90.1 & 77.7 & 75.3 & 73.8 & 75.6 & 73.9 & 73.9 & 73.6 & 72.2 & 71.4 & 75.0 & 67.2 & 69.0 & 59.7 & 73.5 \\
19.2M & 90.0 & 77.4 & 75.2 & 73.6 & 75.7 & 74.4 & 74.1 & 73.8 & 72.5 & 71.3 & 75.1 & 67.9 & 69.1 & 59.7 & 73.6 \\
25.6M & 90.0 & 77.5 & 75.3 & 73.6 & 75.7 & 74.1 & 74.2 & 73.9 & 72.7 & 71.4 & 75.3 & 67.8 & 69.4 & 59.7 & 73.6 \\ \midrule
\multicolumn{1}{c}{}&\multicolumn{14}{c}{\textbf{R\at1000}}& \\
 3.2M & 96.3 & 88.7 & 87.5 & 86.3 & 87.4 & 86.2 & 85.5 & 85.7 & 84.7 & 83.8 & 86.8 & 81.5 & 81.4 & 75.1 & 85.5 \\
 6.4M & 96.5 & 88.4 & 87.3 & 86.1 & 87.1 & 86.7 & 86.0 & 85.7 & 84.8 & 83.6 & 86.8 & 81.6 & 82.2 & 74.8 & 85.5 \\
12.8M & 96.5 & 88.0 & 87.5 & 85.8 & 86.9 & 86.0 & 85.4 & 85.6 & 84.7 & 83.6 & 86.4 & 80.9 & 81.4 & 74.3 & 85.2 \\
19.2M & 96.6 & 87.8 & 87.2 & 85.9 & 86.8 & 86.5 & 85.2 & 85.4 & 84.4 & 83.2 & 86.9 & 81.1 & 81.5 & 74.0 & 85.2 \\
25.6M & 96.7 & 87.8 & 87.3 & 85.8 & 87.0 & 86.2 & 85.3 & 85.4 & 84.3 & 83.4 & 87.0 & 80.9 & 81.6 & 74.0 & 85.2 \\
\bottomrule
\end{tabular}
}
\caption{Influence of training samples on the performance of ColBERT-XM model on mMARCO small dev set.}
\label{tab:training_samples}
\end{table*}

\section{Centroid-based Indexing\label{app:indexing}}
ColBERTv2's centroid-based indexing consists of three main stages \citep{santhanam2022colbertv2}. 

First, we select a set of cluster centroids $\mathcal{C} = \{\mathbf{c}_j \in \mathbb{R}^{d_{\mathrm{out}}}\}_{j=1}^{N}$ of size $N$, proportional to the square root of the estimated number of term embeddings across the entire passage collection, by applying $k$-means clustering to the contextualized term embeddings of only a sample of all passages.

Then, every passage in the corpus is processed using the modular language representation model, as detailed in \Cref{subsec:xmod}, and the resulting contextualized term embeddings are assigned the identifier of the closest centroid $\mathbf{c}_j \in \mathcal{C}$, which requires $\lceil\log_2 |\mathcal{C}|\rceil$ bits to be encoded. Additionally, a \textit{residual} representation $\mathbf{r}^{p}_i \in \mathbb{R}^{d_{\mathrm{out}}}$ is computed for each term embedding to facilitate its reconstruction given $\mathbf{r}^{p}_i=\mathbf{\hat{h}}^{p}_i-\mathbf{c}_j$. To enhance storage efficiency, each dimension of this residual vector is quantized into 2-bit values. Consequently, storing each term vector requires $2d_{\mathrm{out}}+\lceil\log_2 |\mathcal{C}|\rceil$ bits, i.e. roughly 7$\times$ less than the $16d_{\mathrm{out}}$ bits needed for the 16-bit precision compression used in ColBERTv1, without compromising on retrieval quality.

Finally, the identifiers of the compressed term embeddings linked to each centroid are grouped together and saved on disk within an inverted list. At search time, the $n_{\mathrm{probe}}$ centroids closest to every term representation of a given query are identified, and the embeddings indexed under these centroids are fetched for a first-stage candidate generation. Specifically, the compressed embeddings associated with the selected centroids are accessed via the inverted list structure, decompressed, and scored against each query vector using the similarity metric. The computed similarities are then aggregated by passage for each query term and subjected to a max-pooling operation. Since not all terms from a given passage are evaluated but only those associated with the selected centroids, the scores from this preliminary retrieval stage serve as an approximate of the MaxSim operation described in \Cref{subsec:maxsim}, thus providing a lower bound on actual scores. These approximated values are summed across query terms, and the $k$ passages with the highest scores undergo a secondary ranking phase. Here, the full set of term embeddings for each candidate passage is considered to calculate the exact MaxSim scores. The selected passages are then reordered based on these refined scores and returned.

\section{Experimental Details \label{app:experiment_details}}
\Cref{tab:training_samples} provides a comprehensive breakdown of ColBERT-XM's performance across individual languages on mMARCO small dev set, depending on the number of examples used for training.

\section{Reproducibility \label{app:reproducibility}}
We ensure the reproducibility of the experimental results by releasing our code on Github at {\small \url{https://github.com/ant-louis/xm-retrievers}}. In addition, we release our model checkpoints on Hugging Face at {\small \url{https://huggingface.co/antoinelouis/colbert-xm}} and {\small \url{https://huggingface.co/antoinelouis/dpr-xm}}.

\end{document}